%% file: main.tex
\documentclass{article}

\PassOptionsToPackage{numbers, compress}{natbib}

\usepackage[main, final]{neurips_2026}

\input{preamble}

\usepackage[utf8]{inputenc}
\usepackage[T1]{fontenc}
\usepackage{listings}
\usepackage[
    colorlinks=true,
    linkcolor=PaperLink,
    citecolor=blue,
    urlcolor=PaperBlue
]{hyperref}
\usepackage{url}
\usepackage{amsfonts}
\usepackage{microtype}

\title{
Lang-SVG: Hierarchical Image Vectorization with Language Priors}

\author{
  \textbf{Xi Liu}$^{1}$ \quad
  \textbf{Chaoyi Zhou}$^{1}$ \quad
  \textbf{Run Wang}$^{1}$ \quad
  \textbf{Jiaang Li}$^{2}$ \\
  \textbf{Feng Luo}$^{1}$ \quad
  \textbf{Junxiang Huang}$^{3}$ \quad
  \textbf{Siyu Huang}$^{1}$\thanks{\emph{Corresponding author: Siyu Huang}}\\ \vspace{-1em} \\
  $^{1}$Clemson University \quad
  $^{2}$University of Copenhagen \quad
  $^{3}$eBay
}

\begin{document}

\maketitle
\input{figs/teaser/item}
\input{sec/0_abstract}

\input{sec/1_introduction}
\input{sec/2_related_work}

\input{sec/3_method}
\input{sec/4_experiments}

\input{sec/5_conclusion}

\input{sec/6_acknowledgments}

\newpage
\bibliographystyle{unsrt}
\bibliography{neurips_2026}

\appendix
\input{sec/7_appendix}

\clearpage

\newpage

\end{document}

%% file: preamble.tex
\usepackage[table, dvipsnames]{xcolor}
\input{paper_colors}

\usepackage{booktabs}
\usepackage{multirow}
\usepackage{graphicx}

\usepackage{amsmath}
\usepackage{amssymb}
\usepackage{bm}
\usepackage{etoolbox}
\usepackage{algorithm}
\usepackage{algpseudocode}

\AtEndPreamble{
    \usepackage[capitalize]{cleveref}
}
\makeatletter
\AtBeginDocument{%
    \providecommand*{\theHALG@line}{}%
    \renewcommand*{\theHALG@line}{\thealgorithm.\arabic{ALG@line}}%
}
\makeatother

\newcommand{\cmark}{\checkmark}

\newcommand{\ADE}{ADE20K~\cite{zhou2019semantic}}
\newcommand{\LIVE}{LIVE~\cite{ma2022towards}}
\newcommand{\LIVSS}{LIVSS~\cite{wang2025layered}}
\newcommand{\DiffVG}{DiffVG~\cite{Li:2020:DVG}}
\newcommand{\Ours}{Lang-SVG}

\newcommand{\image}{\mathbf{I}}
\newcommand{\svgtree}{\mathcal{T}}
\newcommand{\svglayer}[1]{\mathcal{L}_{#1}}
\newcommand{\node}[1]{v_{#1}}

\newcommand{\realset}{\mathbb{R}}
\newcommand{\imgheight}{H}
\newcommand{\imgwidth}{W}

\newcommand{\nodeset}{\mathcal{V}}
\newcommand{\edgeset}{\mathcal{E}}
\newcommand{\rootnode}{r}
\newcommand{\numnodes}{K}

\newcommand{\nodeidx}{k}
\newcommand{\childidx}{k'}

\newcommand{\semtag}[1]{y_{#1}}
\newcommand{\mask}[1]{\bm{m}_{#1}}
\newcommand{\layermasks}[1]{\mathcal{M}^{(#1)}}
\newcommand{\numlayers}{L}
\newcommand{\level}{\ell}

\renewcommand{\paragraph}[1]{\vspace{.4em}\noindent\textbf{#1}}

%% file: paper_colors.tex
\definecolor{PaperBlue}{HTML}{2B5F8C}
\definecolor{PaperLink}{HTML}{37647F}
\definecolor{PaperBest}{HTML}{DDEBF7}
\definecolor{PaperRule}{HTML}{CAD7E2}
\definecolor{PaperCanvas}{HTML}{F7F9FB}
\definecolor{PaperWarm}{HTML}{946A3A}
\definecolor{PaperMuted}{HTML}{6B7280}

\colorlet{1st}{PaperBest}
\colorlet{oursblue}{PaperBest}

%% file: figs/teaser/item.tex
\vspace{-1.0em}

\noindent
\includegraphics[width=\textwidth]{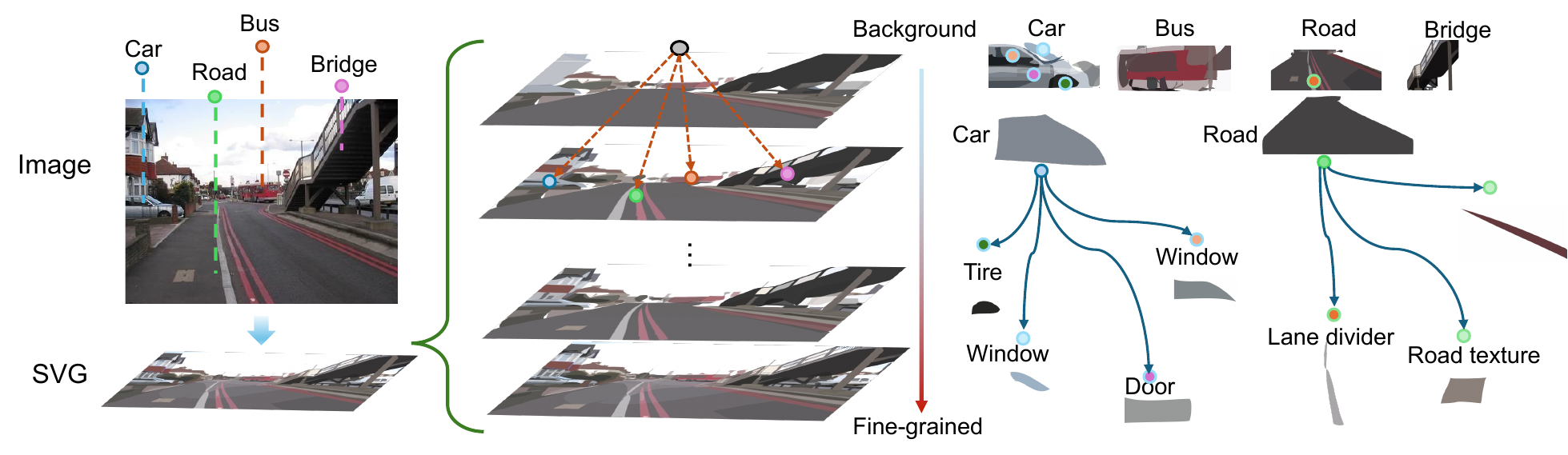}

\refstepcounter{figure}
\label{fig:teaser}
{\small
\textbf{Figure~\thefigure:}
This work proposes \textbf{\Ours}, a tree-structured SVG representation that vectorizes natural images into hierarchical object-part SVGs with language labels.
}

%% file: sec/0_abstract.tex
\vspace{0.7em}
\begin{abstract}
\vspace{-0.3em}
Image vectorization reconstructs raster images as compact scalable vector graphics (SVG), providing a structured and editable code representation of images. However, most existing vectorization methods optimize for pixel-level rendering fidelity, often producing SVGs that lack alignment with human-perceived object-part hierarchies, making them difficult to manipulate. To address this, we propose Lang-SVG, which introduces an explicit tree structure to organize SVG primitives with semantic labels and parent-child relationships. Lang-SVG incorporates granularity-controllable and open-vocabulary segmentation priors to initialize object-part hierarchies and leverages multimodal large language models (MLLMs) with SVG tree context to assign semantic labels to unlabeled nodes. To reduce redundancy, we introduce depth-adaptive simplification to reduce shape complexity and Tree-Guided Pruning and Merging to eliminate redundant SVG nodes. L1-SAM-Prompting further recovers missing regions to improve reconstruction fidelity and structural completeness. Beyond reconstruction fidelity, we introduce a new evaluation protocol that measures structural quality and semantic part alignment. Experiments demonstrate that Lang-SVG achieves state-of-the-art performance in both rendering fidelity and structural quality.
Project website: \url{https://xiliu8006.github.io/Lang-SVG/}.
\end{abstract}

%% file: sec/1_introduction.tex
\vspace{-0.5em}
\section{Introduction}
\label{sec:introduction}
Image vectorization transforms raster images into compact scalable vector graphics (SVG), providing a structured, renderable, and editable code representation of visual content. As a text-based representation, SVG is increasingly becoming an interface between images and language models. Recent MLLMs learn to generate SVG code~\cite{rodriguez2025starvector,yang2025omnisvg,wu2024chat2svg,yun2026vector}, benchmarks evaluate visual understanding through image-to-SVG generation~\cite{lin2025vcode,chen2025svgenius}, and code-based approaches~\cite{gao20263dcodeverse,kundu2026physimcode,wang2026code,kaduri2026reinforcement,ye2026genclaw} explore structured visual code for generation, manipulation, and reasoning. However, existing SVG datasets ~\cite{rodriguez2025starvector,yang2025omnisvg, wang2025svgen, iconfont} largely focus on icons and illustrations, offering limited coverage of complex natural images. Vectorizing natural images offers a much larger and more diverse source of SVG code beyond simple icons and illustrations, but representing their complex visual content as structured code remains challenging.

Recent differentiable SVG methods~\cite{Li:2020:DVG,liu2026bzier,11094824,hirschorn2024optimize} have substantially advanced image vectorization in terms of rendering fidelity, efficiency, and compactness, with recent approaches~\cite{wang2025layered,zhou2024segmentation,ma2022towards,yun2026vector} further exploring layered and semantic representations to improve editability. However, these methods typically organize SVG primitives as flat layers without explicitly encoding object-part hierarchies. As a result, primitives belonging to the same object may be distributed across different layers, making it difficult to select and edit individual objects or their parts. Moreover, without hierarchical guidance, SVG optimization may allocate excessive primitives to redundant shapes or high-frequency textures rather than meaningful semantic regions. These limitations highlight the need for image vectorization that preserves semantic structure while maintaining high rendering fidelity.

We argue that this limitation partly stems from the lack of explicit semantic and structural constraints in existing image vectorization objectives. While pixel-level reconstruction losses effectively optimize SVG geometry, opacity, and color, they do not explicitly encourage primitives to form meaningful semantic structures. SGLIVE ~\cite{zhou2024segmentation} incorporates SAM~\cite{kirillov2023segment} to improve structural alignment, but SAM masks do not explicitly capture fine-grained semantic object-part hierarchies.
\LIVSS~ further employs diffusion-based simplification to generate images at multiple semantic levels, followed by SAM to extract masks from coarse semantic level to fine-grained level. However, diffusion-based simplification may distort geometric details, causing shape boundaries to deviate from the image.

To address these issues, we propose \textbf{\Ours}, a tree-structured SVG representation that organizes vector primitives with semantic labels and parent-child hierarchy, as illustrated in Fig.~\ref{fig:teaser}.
Unlike flat SVG representations, our hierarchical structure explicitly models object-part relationships while preserving the flexibility of free-form vector primitives.
We initialize its semantic hierarchy with open-vocabulary segmentation~\cite{ren2024grounded} and enrich it with coarse-to-fine masks from granularity-controllable models such as Semantic-SAM~\cite{li2024segment}, avoiding the geometric distortions introduced by diffusion-based simplification.
However, these masks often contain overlapping or redundant regions, making direct vectorization difficult under a limited control-point budget.
We therefore introduce depth-adaptive simplification to allocate geometric complexity according to tree depth, and a Tree-Guided Pruning and Merging module to remove redundant regions while preserving semantically labeled nodes.
We further introduce L1-SAM-Prompting, which uses reconstruction residuals to prompt SAM for coherent missing regions, progressively refining the tree beyond the initial segmentation masks.
Our experiments demonstrate that the explicit tree structure improves both reconstruction fidelity and alignment with human-perceived image structures.

Beyond hierarchical image vectorization, Lang-SVG further supports SVG semantic labeling. While open-vocabulary segmentation provides initial semantic labels, finer regions introduced during tree refinement may remain unlabeled. To address this, we develop a tree-guided semantic tagging method that leverages multimodal large language models (MLLMs) with visual evidence and hierarchical SVG context, including parent, sibling, and child relationships. This enables semantic labeling beyond the initial segmentation models, supporting semantic selection and part-level manipulation.

Evaluating hierarchical SVG vectorization requires going beyond photometric and perceptual metrics, which cannot determine whether SVG primitives align with human-perceived object-part structures. We therefore introduce a new evaluation protocol that combines StructXLIP~\cite{ruan2026StructXLIP}, a structure-aware vision-language model for measuring image-level structural similarity, with part/subpart recall to assess the alignment between SVG regions and human annotations at different IoU thresholds. On the \ADE~dataset, Lang-SVG improves StructXLIP from 0.462 to 0.506 and average part recall from 28.8\% to 48.5\% at 576 control points, while achieving the best reconstruction performance among the compared methods. Furthermore, incorporating SVG tree context improves semantic tagging accuracy from 58.5\% to 81.3\% in our masked-label evaluation, demonstrating the value of hierarchical information for semantic reasoning.

Overall, our contributions are summarized as follows:
\begin{itemize}
    \item  To the best of our knowledge, this work is the first to construct hierarchical SVG trees with semantic labeling from raster images.

     \item We propose \Ours, a novel hierarchical SVG representation with semantic labels and parent-child relationships. To handle the complex and redundant masks introduced by Semantic-SAM's fine-grained, multi-level segmentation, we introduce depth-adaptive simplification and tree-guided pruning and merging. We further introduce L1-SAM-Prompting to recover missing regions and tree-aware MLLM labeling for semantic tagging.

    \item We propose a new evaluation protocol for image vectorization that goes beyond reconstruction fidelity by measuring structural quality and semantic part alignment. The experiments show that \Ours~achieves state-of-the-art results under both reconstruction fidelity and structural quality.
\end{itemize}

%% file: sec/2_related_work.tex
\section{Related Work}
\label{sec:related work}

\paragraph{Image vectorization.}
Image vectorization transforms a raster image into a parametric representation, enabling resolution-independent rendering while making the image directly editable by humans~\cite{reddy2021im2vec, lopes2019learned, Cao_2023_CVPR, zhang2024text, hu2024supersvg}. DiffVG~\cite{Li:2020:DVG} was among the first works to make the rendering of SVG differentiable, reformulating the vectorization task as an end-to-end optimization problem. Bézier Splatting~\cite{liu2026bzier} improves rendering efficiency by introducing a splatting-based formulation inspired by 2D Gaussian representations~\cite{zhang2024gaussianimage}, making vectorization more practical for high-resolution rendering. \LIVE~introduces a layer-wise coarse-to-fine strategy that progressively adds primitives, improving the quality and topology of SVG representations. Subsequent works~\cite{hirschorn2024optimize, 11094824} address the over-complexity of SVGs by simplifying vector representations and applying add-and-prune strategies to remove redundant primitives. Beyond compactness, SGLIVE~\cite{zhou2024segmentation}, \LIVSS{} and SemLayer~\cite{Xu_2026_CVPR} incorporate segmentation priors to better align vectorized shapes with the structural information underlying natural image textures, leading to more semantically meaningful decompositions. AmodalSVG~\cite{hu2026amodalsvg} goes further by peeling semantically complete raster layers with MLLM guidance and inpainting.

A recent line of work reformulates vectorization as an SVG code generation task with MLLMs. StarVector~\cite{rodriguez2025starvector}, OmniSVG~\cite{yang2025omnisvg}, InternSVG~\cite{wang2025internsvg} and  IR4RL~\cite{kaduri2026reinforcement} perform image-to-SVG generation directly in the SVG code space, leveraging semantic visual understanding to predict richer primitives and more compact outputs.
These methods indicate that vectorization is moving from pure reconstruction toward editability and semantic decomposition.
However, most existing methods still represent SVGs as an ordered set of primitives, or first separate semantic regions in the raster domain and then vectorize them independently. In contrast, our proposed \Ours~representation explicitly encodes parent-child relations and semantics among SVG nodes, enabling applications such as grouping and semantic analysis of SVG primitives.

\paragraph{Open-vocabulary segmentation.}
Open-vocabulary segmentation provides a flexible way to extract semantic regions from images using language-level concepts, which is particularly useful for semantic vectorization. The SAM family~\cite{kirillov2023segment, ravi2025sam, carion2026sam} has enabled general-purpose promptable mask generation across diverse image domains, with recent variants further extending prompts from points and boxes to concept-level inputs. However, these semantic prompts are typically provided by users. Grounded SAM~\cite{ren2024grounded} addresses this limitation by coupling open-vocabulary detection or tagging with SAM-based segmentation: semantic candidates are first localized or generated automatically, and SAM is then applied to produce the corresponding masks. In this way, language-level concepts can be converted into raster-domain candidate masks with less manual prompting, providing useful guidance for downstream vectorization. Some works~\cite{li2024partglee,miao2026langhops} further target open-vocabulary part segmentation and object-part parsing. Another line of work~\cite{li2024segment,wang2024segment,yu2025unsamv2} improves the granularity and semantic controllability of segmentation foundation models, enabling finer-grained masks. Our work leverages open-vocabulary segmentation and granularity-controllable SAM models to introduce semantically aligned hierarchical information into image vectorization. This hierarchical structure allows vectorized layers to better align with object-part semantics, and further supports tagging semantically meaningful regions that are not explicitly parsed by the segmentation model.

\paragraph{Multimodal large language model.}
Multimodal large language models (MLLMs) are typically pretrained via next-token prediction over heterogeneous inputs, including images, text, and videos. Early MLLMs, such as BLIP~\cite{li2022blip}, Flamingo~\cite{alayrac2022flamingo}, and Idefics~\cite{laurenccon2023obelics}, introduce cross-attention mechanisms to facilitate interaction between visual and linguistic representations. Modern MLLMs largely follow the paradigm popularized by LLaVA~\cite{liu2023visual}, where visual tokens produced by pretrained vision encoders such as CLIP~\cite{radford2021learning} are projected into the LLM backbone. This design simplifies multimodal fusion while retaining strong language modeling capabilities. Recent research shows that MLLMs can ground language in visual content, leveraging both visual and text context with implicit commonsense knowledge acquired during large-scale training~\cite{deitke2025molmo,wu2024v,comanici2025gemini,achiam2023gpt,li2026ravenea}. We utilize MLLMs to predict unknown tags of SVG representations conditioned on visual inputs and structured semantic context derived from a tree-based representation.

%% file: sec/3_method.tex
\section{Methodology}
\label{sec:method}

Given an input raster image $\image \in \realset^{\imgheight \times \imgwidth \times 3}$, we aim to construct a tree-structured SVG representation $\svgtree$ that achieves high-fidelity rendering while preserving the semantic hierarchy and fine-grained structural details of the original image. We define
\begin{equation}
    \svgtree = (\nodeset, \edgeset, \rootnode),
    \qquad
    \node{\nodeidx} = (\semtag{\nodeidx}, s_{\nodeidx}, \mask{\nodeidx}).
\end{equation}
Here, $\nodeset=\{\node{\nodeidx}\}_{\nodeidx=1}^{\numnodes}$ is the set of nodes, $\edgeset$ defines the parent--child relationships, and $\rootnode$ is the root node. Each edge $(\node{\nodeidx}, \node{\childidx}) \in \edgeset$ indicates that $\node{\childidx}$ is a semantic constituent of $\node{\nodeidx}$. Each node $\node{\nodeidx}$ stores a semantic tag $\semtag{\nodeidx}$, a semantic mask $\mask{\nodeidx}$, and a set of SVG primitives parameterized by $s_{\nodeidx}$.

\subsection{SVG Tree Construction via Granularity-Controllable Segmentation}
\label{sec:tree_construction}

To construct a tree-structured SVG representation that preserves semantic hierarchies and fine-grained structural details, we first obtain masks with semantic tags from open-vocabulary~\cite{ren2024grounded} or subpart-aware segmentation methods~\cite{miao2026langhops,li2024partglee}, as shown in Fig.~\ref{fig:method}. These masks initialize a semantic tree with labeled object-part regions. We then enrich the tree with multi-granularity masks from granularity-controllable segmentation models, such as Semantic-SAM~\cite{li2024segment} and UnSAM~\cite{wang2024segment}. These additional masks are initially assigned \emph{unknown} semantic tags and serve as candidate nodes for refining the hierarchy. Formally, the multi-granularity masks are defined as
\begin{equation}
    \{\layermasks{\level}\}_{\level=1}^{\numlayers},
    \qquad
    \layermasks{\level} = \{\mask{i}^{(\level)}\},
\end{equation}
where each set $\layermasks{\level}$ contains masks at a specific semantic granularity.

We construct the hierarchy by inserting candidate masks from coarse to fine based on spatial containment. A node $\node{\nodeidx}$ is considered a candidate parent of $\node{\childidx}$ if
\begin{equation}
    \frac{|\mask{\childidx} \cap \mask{\nodeidx}|}
    {|\mask{\childidx}|} \geq \tau.
\end{equation}
Among valid parents, we select the node at the greatest tree depth, breaking ties by mask IoU. If no valid parent exists, the node is attached to the full-image root. This process organizes segmentation masks into explicit parent-child relationships rather than flat layers, providing a hierarchical structure for subsequent SVG initialization and optimization. Appendix~\ref{sec:appendix_algorithms} provides the complete construction and update procedures.
\input{figs/method/item}

\subsection{Tree-Guided SVG Optimization}
\label{sec:svg_optimization}

\paragraph{Tree-guided SVG initialization.} For each node $\node{\nodeidx}$, we convert its mask $\mask{\nodeidx}$ into a polygon, which is then approximated by a set of B\'{e}zier curves to initialize the corresponding vector layer $\svglayer{\nodeidx}$ as a closed region. A key challenge arises during this conversion: masks from coarser segmentation levels tend to have complex, irregular boundaries, yet due to alpha compositing, nodes closer to the tree root are more occluded and contribute less to the final rendered image. Fitting all masks with high-complexity polygons therefore introduces redundant primitives with little gain in reconstruction quality. To address this, we propose depth-adaptive simplification. Rather than using a fixed tolerance for all nodes, we adapt the approximation tolerance according to the node's depth in the tree: nodes closer to the root are fitted with higher tolerance, yielding simpler polygons, while leaf nodes use lower tolerance to preserve fine-grained boundary details. This allocates geometric complexity in proportion to visual contribution, achieving a better balance between reconstruction fidelity and primitive efficiency.

\paragraph{L1-SAM-Prompting.}
A common practice in image vectorization is to initialize SVG primitives from clusters derived from the L1 loss map between the current rendering and the target image. While this improves reconstruction metrics such as PSNR, it tends to allocate primitives toward fitting high-frequency textures rather than capturing structurally meaningful regions, resulting in insufficient representation of semantic structure. Although \LIVSS~incorporates SAM to generate masks, it still relies on loss-driven clustering to meet path budgets and improve visual quality, since even with dense point prompts, SAM is not intended to produce fine-grained masks.

To address this, we propose L1-SAM-Prompting, which bridges loss-driven mask generation and SAM. Instead of directly using L1 clustering masks for SVG initialization, we extract their bounding boxes and mean points as prompts for SAM, producing masks that are more coherent and structurally complete, rather than dominated by frequency discrepancies between low-frequency SVG renderings and high-frequency natural images.
We integrate this strategy as an intermediate step between two optimization stages: a \emph{structure optimization} stage that fits segmentation-derived masks, and a \emph{visual refinement} stage that refines appearance details. The L1-SAM-Prompting step generates supplementary masks that emphasize meaningful structures rather than noise from pixel-level loss, improving the balance between structural representation and visual fidelity.

\paragraph{Tree-guided pruning and merging.}
Multi-granularity segmentation produces a large number of masks that are often redundant and unstable. A common approach is to prune masks based on spatial overlap (e.g., IoU); however, IoU-based pruning is insufficient, as it fails to remove masks with different shapes but similar appearance and semantics. Rather than relying solely on geometric overlap, our representation enables appearance-aware pruning using the associated SVG layers. Specifically, we compare each node with its parent based on the color of their SVG layers: if an unlabeled child node exhibits high color similarity to its parent, we remove it; parent--child or sibling nodes within the same semantic branch can be merged when they exhibit similar color and sufficient spatial overlap. These branch constraints prevent similarly colored regions from different objects from being merged. Color and overlap therefore identify rendering redundancy within a branch rather than establish semantic identity. Nodes that already carry semantic tags are preserved and excluded from pruning or merging to retain semantic information.

\subsection{Tree-Guided Semantic Labeling with Multimodal Large Language Models}
\label{sec:semantic_assignment}
Given a tree-structured SVG representation $\svgtree$, our objective is to assign semantic labels to nodes with unknown tags. We focus on unlabeled nodes within subtrees rooted at semantically labeled nodes, allowing us to exploit hierarchical context. As shown in Fig.~\ref{fig:tagging}, we use Qwen3.5-27B~\cite{yang2025qwen3} to predict the semantic label of each unlabeled node by jointly considering its local neighborhood and ancestral context within the tree.

For each unlabeled node $\node{\nodeidx}$, we construct a structured prompt containing the full image, a crop extracted from the bounding box of the target SVG region, and its RGB color from SVG primitives. We further include parent and ancestor tags to define the semantic scope, along with labeled sibling and child tags and sibling bounding boxes to provide local relational and spatial cues for disambiguation. The prompt is formulated as a structured decision task, where the MLLM is asked to infer the semantic tag of the target node based on these visual and hierarchical cues. We consider both closed-set and open-vocabulary settings. In the closed-set setting, the MLLM selects a label from a predefined candidate set, while in the open-vocabulary setting, it predicts a free-form tag. Ambiguous nodes are assigned a default \emph{texture} or \emph{unknown} tag to avoid unreliable semantic over-assignment. This process augments the SVG tree with node-level semantic labels, enabling vector primitives to be organized by both geometry and semantic identity. Table~\ref{tab:qwen-subpart-fair-ablation} shows that the tree structure provides useful context for semantic tag recovery.

%% file: figs/method/item.tex
\begin{figure*}
  \centering
  \includegraphics[width=\linewidth]{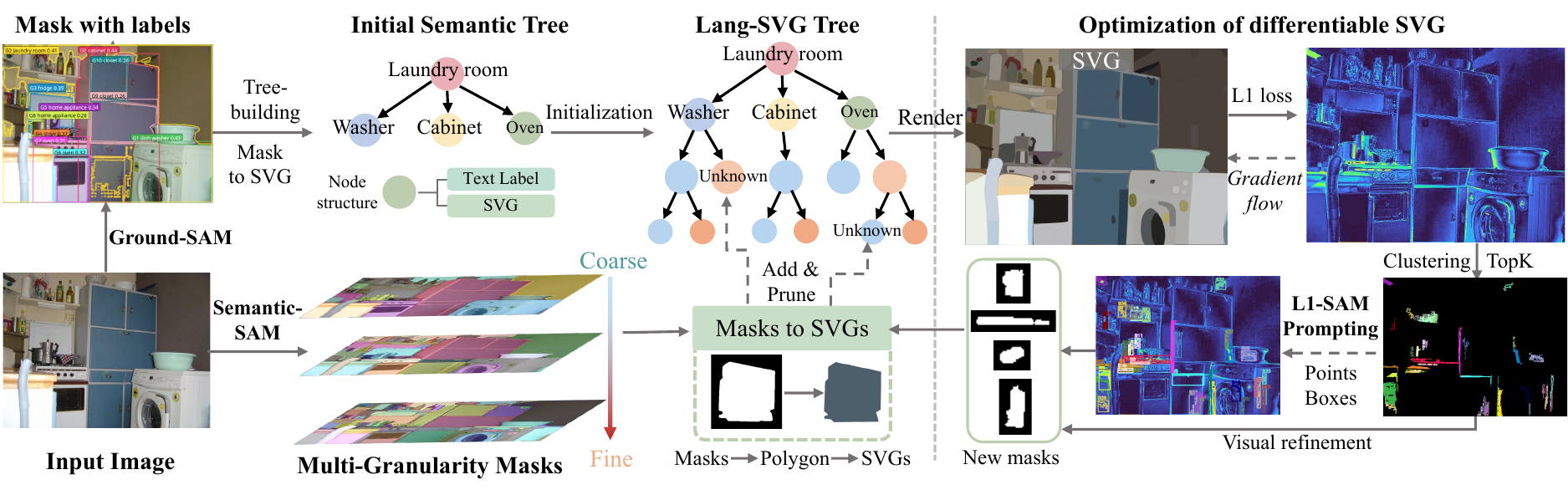}
  \\
  \vspace{-0.2em}
\caption{
\textbf{Overview of Lang-SVG framework.}
Given an input image, we initialize a hierarchical Lang-SVG tree from multi-granularity and open-vocabulary segmentation masks. The tree is iteratively refined through differentiable SVG optimization, L1-SAM-Prompting to recover missing regions, and Add \& Prune operations to add new leaves and prune redundant ones.
}
\label{fig:method}
\vspace{-0.5em}
\end{figure*}

%% file: sec/4_experiments.tex
\vspace{-.2em}
\section{Experiments}
\vspace{-.2em}
\label{sec:experiments}

\subsection{Experiment Setup}
\label{sec:exp_setup}

\paragraph{Datasets.}
Existing image vectorization methods ~\cite{ma2022towards, zhou2024segmentation}, support only reconstruction metrics and provide neither multi-level object-part hierarchies nor node-level semantic labels. Therefore, we therefore conduct our main evaluation on ADE20K-Part234~\cite{zhou2019semantic}, whose diverse natural scenes are annotated with object, part, and subpart masks and labels. These annotations allow us to test whether SVG regions align with human-annotated parts, whether parent-child relations and semantic tags are correct, and whether tree context helps recover missing tags. We uniformly sample 200 validation images, which contain 752 ground-truth part/subpart masks. To test generalization, we further use COCO-AOKVQA~\cite{schwenk2022okvqa}, whose image descriptions enable text-to-image semantic evaluation, and 40 complex SVG-style images selected from Iconfont \cite{iconfont}.

\paragraph{Evaluation metrics.}
We first evaluate reconstruction fidelity using SSIM, PSNR, LPIPS, and CLIP similarity. However, these metrics do not fully capture the structural details of an image or its semantic object-part hierarchy. Therefore, we adopt StructXLIP~\cite{ruan2026StructXLIP}, a structure-aware vision-language model trained with edge-based cues, to measure image-level structural similarity, and introduce part/subpart recall on ADE20K-Part234 to evaluate the alignment between SVG regions and ground-truth parts and subparts at different IoU thresholds. To evaluate semantic alignment, we compute text-to-image ($t2i$) CLIP and StructXLIP similarities between COCO-AOKVQA image descriptions and rendered SVGs.

To evaluate the correctness of SVG semantic hierarchies, we introduce three additional metrics: \emph{Child Containment} measures the fraction of an SVG child's area covered by its ground-truth parent; \emph{Tagging Reliability} measures the fraction of a labeled SVG region covered by the ground-truth subpart with the same label; and \emph{Labeled Subpart Recall} measures the fraction of ground-truth subparts recovered by SVG children with the correct parent, label, and IoU above a given threshold. All three metrics are reported at thresholds of 0.3, 0.5, and 0.7.

\paragraph{Evaluation mode.}
Since different vectorization methods use different SVG parameterizations, a fair comparison should control the total number of points. \LIVE~and \DiffVG~use a fixed representation, so the number of paths directly determines the total point budget. In contrast, \LIVSS~and our method convert masks to SVG paths, and each path may contain a different number of points. Therefore, we report two protocols. \textit{Same points} is our main protocol, where all methods are compared under the same total number of points. We also report \textit{Same paths} for \LIVSS~and our method, using the same number of SVG paths to ensure a fair comparison.

\subsection{Comparison to Baselines}
\paragraph{Quantitative and Qualitative results.} We compare Lang-SVG with state-of-the-art differentiable SVG
vectorization methods, including SGLIVE~\cite{zhou2024segmentation}
and LIVSS~\cite{wang2025layered}, which also leverage segmentation priors.
As shown in Tables~\ref{tab:ade20k_point} and~\ref{tab:ade20kpart_recall},
our method consistently outperforms existing methods in both
reconstruction fidelity and structural quality across all point budgets.
At 576 control points, Lang-SVG improves StructXLIP from the
second-best 0.462 to 0.506, and average part recall from 28.8 to 48.5.
The improvements also hold under the \textit{Same paths} protocol.
As shown in Fig.~\ref{fig:main-comparison}, our method better preserves
coherent object regions and fine-grained structures, especially under
limited primitive budgets, rather than fitting local textures or
reconstruction residuals.
These results demonstrate that Lang-SVG achieves better visual fidelity
and object-part structural alignment with a limited primitive budget.

\input{tbls/ade20k_comparison/item}
\input{tbls/part_recall/item}

\paragraph{Cross-dataset generalization.}
To further evaluate generalization, we test image vectorization on COCO-AOKVQA~\cite{schwenk2022okvqa} and Iconfont~\cite{iconfont}. COCO-AOKVQA provides scene descriptions, allowing us to evaluate semantic alignment between the text descriptions and rendered SVG images using text-to-image CLIP and StructXLIP scores. As shown in Table~\ref{tab:coco_point}, our method achieves the best reconstruction and semantic scores across all point budgets. At 576 points, compared with DiffVG, the strongest PSNR baseline, PSNR improves modestly from $20.35$ to $20.53$, while text-to-image StructXLIP increases substantially from $0.188$ to $0.270$, indicating better preservation of semantic content beyond pixel-level reconstruction. On Iconfont SVG-style images, our method also achieves the highest PSNR, improving from $28.62$ (SGLIVE) to $29.28$. Additional results are provided in Appendix~\ref{sec:appendix_iconfont}.
\input{tbls/coco_comparison/item}

\input{figs/comparison/item}

\subsection{Evaluation of Hierarchy and Semantic Quality}
\vspace{-0.2em}
\label{sec:hierarchy_quality}
Table~\ref{tab:hierarchy_quality} evaluates the hierarchy and semantic quality of the generated SVG trees. Child Containment reaches 95.55\% at threshold 0.5 and remains 86.65\% at 0.7, demonstrating reliable parent-child relationships. At threshold 0.5, Tagging Reliability reaches 68.25\%, while Labeled Subpart Recall reaches 47.06\%, indicating that our method recovers meaningful semantic labels and object-part structures. Since semantic tags and subpart coverage depend partly on upstream segmentation, stronger segmentation models could further improve these results.

\input{tbls/hierarchy_quality/item}

\subsection{Semantic SVG Tagging}
\label{sec:qwen_tag_assignment}

To evaluate whether the hierarchical structure of \Ours~provides useful semantic information beyond reconstruction, we conduct a node-wise tag recovery task. Starting from ADE20K ground-truth masks and labels, we construct SVG trees following Sec.~\ref{sec:tree_construction} and randomly hide up to $50\%$ of the part and subpart tags in each scene, while preserving the masks, node locations, and tree topology. The hidden tags are used only for evaluation. Given the full image, target region, and different subsets of tree context, Qwen3.5-27B predicts the missing tags from the closed ADE20K label set. This setup isolates semantic tagging from segmentation quality and evaluates the contribution of hierarchical context to label recovery.

\input{tbls/tagging_ablation/item}
Table~\ref{tab:qwen-subpart-fair-ablation} shows that hierarchical context substantially improves tag prediction for subpart nodes. Visual evidence alone achieves $58.5\%$ exact accuracy, which increases to $67.7\%$ with object and parent identity and $68.9\%$ with additional sibling and child labels. Including the geometric layout of parent, sibling, and child regions further improves accuracy to $80.2\%$, while adding region color statistics raises it to $81.3\%$. We further examine the contribution of visual inputs in A5 and A6 by removing localized crops or all image inputs. Even without these inputs, the structured context achieves accuracy close to the visual-only baseline, suggesting that the SVG tree itself provides substantial semantic information. These results demonstrate the effectiveness of hierarchical SVG context for node-level semantic tag recovery.

\input{figs/tagging/item}
We further evaluate the full pipeline without ground-truth masks or labels. In this setting, \Ours~is initialized using open-vocabulary part segmentation~\cite{miao2026langhops}, followed by tree-guided semantic tagging of unlabeled SVG nodes. As shown in Fig.~\ref{fig:tagging}, qualitative results on Pascal-Part~\cite{chen2014detect} demonstrate that our method can recover meaningful semantic labels for regions not recognized by the pretrained segmentation model. By leveraging hierarchical relationships, spatial layout, and appearance cues, the MLLM can identify missing object-part labels and, in some cases, finer subpart or texture-level regions. Ambiguous nodes can be assigned \emph{texture} or \emph{unknown} tags to avoid unreliable semantic predictions.

\subsection{Ablation Study}
\label{sec:ablation_study}

We conduct three ablation studies on \ADE~under the 2304-point parameter budget to examine segmentation-prior replacement, tree-related designs, and the remaining modules.

\paragraph{Segmentation priors and vectorization design.}
Table~\ref{tab:segmentation_prior_control} compares segmentation-prior replacement in LIVSS with our complete vectorization framework. Replacing SAM with Semantic-SAM and Grounded-SAM in the LIVSS pipeline increases PSNR from 24.12 to 24.23 and StructXLIP from 0.6950 to 0.7081. Using the same segmentation priors, \Ours~further improves PSNR to 25.90 and StructXLIP to 0.775. These results show that the performance gains cannot be explained by stronger segmentation priors alone, supporting the contribution of our vectorization design.
\input{tbls/ablation_priors/item}

\paragraph{Tree structure and tree-guided operations.}
Table~\ref{tab:tree_isolation} evaluates the tree structure, depth-adaptive simplification, and tree-guided pruning and merging using the same segmentation priors. Tree-guided pruning and merging operate on parent--child and sibling relations in the SVG tree and therefore require the tree structure; disabling the tree necessarily disables tree pruning. To evaluate the tree's contribution with pruning disabled, we compare \emph{w/o tree and tree pruning} with \emph{w/o Tree Prune}. Enabling the tree increases PSNR from 24.29 to 25.71 and StructXLIP from 0.7054 to 0.7687. The \emph{w/o depth simplification} variant retains both the tree and tree pruning, reducing PSNR from 25.90 to 24.84 and StructXLIP from 0.775 to 0.7140. Finally, enabling pruning and merging in the \emph{w/o Tree Prune} configuration improves PSNR from 25.71 to 25.90 and StructXLIP from 0.7687 to 0.775. This indicates that redundant masks otherwise remain and compete for the same parameter budget.
\input{tbls/ablation_tree/item}

\paragraph{L1-SAM-Prompting and segmentation backbone.}
As shown in Table~\ref{tab:ablation_v4_2304}, \emph{w/o L1-SAM-Prompting} disables the step that uses the L1 loss map to generate SAM prompts for obtaining additional masks, and instead directly uses the masks from L1 clustering. This variant drops StructXLIP from $0.775$ to $0.762$, because more SVG primitives are allocated to optimizing the pixel-level L1 residual rather than to masks that follow semantic structures. In the \emph{w/ UnSAMv2} variant, we replace Semantic-SAM with UnSAMv2~\cite{yu2025unsamv2} at granularity levels $[0.8, 0.6, 0.4]$, and also replace SAM in L1-SAM-Prompting with UnSAMv2 at granularity $0.1$. Semantic-SAM performs better than UnSAMv2 in our framework. All three ablation tables use the same final \emph{Ours} reference.
\input{tbls/ablation_modules/item}

%% file: tbls/ade20k_comparison/item.tex
\begin{table*}[!t]
\centering
\vspace{-0.15cm}
\caption{\label{tab:ade20k_path}\label{tab:ade20k_point} Quantitative evaluation on ADE20K. Unmarked methods use the \emph{Same points} protocol; $^{\dagger}$ denotes \emph{Same paths}. LIVE and DiffVG use nine control points per path and are reported once for both protocols. S-XLIP denotes StructXLIP.}
\setlength{\tabcolsep}{1.6pt}
\resizebox{0.99\textwidth}{!}{%
\begin{tabular}{l|ccccc|ccccc|ccccc}
\toprule
\textbf{Method} & \multicolumn{5}{c|}{\textbf{576 points / 64 curves}} & \multicolumn{5}{c|}{\textbf{1152 points / 128 curves}} & \multicolumn{5}{c}{\textbf{2304 points / 256 curves}} \\
& $SSIM^{\uparrow}$ & $PSNR^{\uparrow}$ & $LPIPS^{\downarrow}$ & $\text{S-XLIP}^{\uparrow}$ & $CLIP^{\uparrow}$ & $SSIM^{\uparrow}$ & $PSNR^{\uparrow}$ & $LPIPS^{\downarrow}$ & $\text{S-XLIP}^{\uparrow}$ & $CLIP^{\uparrow}$ & $SSIM^{\uparrow}$ & $PSNR^{\uparrow}$ & $LPIPS^{\downarrow}$ & $\text{S-XLIP}^{\uparrow}$ & $CLIP^{\uparrow}$ \\
\midrule
\LIVE & 0.631 & 21.28 & 0.468 & 0.420 & 0.625 & 0.680 & 23.16 & 0.394 & 0.582 & 0.692 & \underline{0.739} & \underline{25.45} & \underline{0.302} & \underline{0.748} & \underline{0.764} \\
\LIVSS & 0.598 & 19.50 & 0.490 & 0.379 & 0.609 & 0.643 & 21.48 & 0.421 & 0.531 & 0.664 & 0.707 & 24.12 & 0.336 & 0.695 & 0.734 \\
\DiffVG & 0.624 & \underline{21.81} & 0.472 & 0.290 & 0.597 & 0.667 & 23.29 & 0.407 & 0.437 & 0.658 & 0.716 & 24.83 & 0.332 & 0.629 & 0.745 \\
SGLIVE~\cite{zhou2024segmentation} & \underline{0.633} & 21.38 & \underline{0.464} & \underline{0.462} & \underline{0.634} & \underline{0.681} & \underline{23.32} & \underline{0.385} & \underline{0.610} & \underline{0.694} & 0.732 & 25.04 & 0.305 & 0.731 & 0.752 \\
\cellcolor{1st} \textbf{Ours} & \cellcolor{1st}\textbf{0.638} & \cellcolor{1st}\textbf{21.88} & \cellcolor{1st}\textbf{0.436} & \cellcolor{1st}\textbf{0.506} & \cellcolor{1st}\textbf{0.651} & \cellcolor{1st}\textbf{0.694} & \cellcolor{1st}\textbf{24.07} & \cellcolor{1st}\textbf{0.360} & \cellcolor{1st}\textbf{0.655} & \cellcolor{1st}\textbf{0.717} & \cellcolor{1st}\textbf{0.746} & \cellcolor{1st}\textbf{25.90} & \cellcolor{1st}\textbf{0.291} & \cellcolor{1st}\textbf{0.775} & \cellcolor{1st}\textbf{0.777} \\
\midrule
LIVSS$^{\dagger}$~\cite{wang2025layered} & \underline{0.629} & \underline{21.13} & \underline{0.445} & \underline{0.500} & \underline{0.652} & \underline{0.668} & \underline{22.68} & \underline{0.387} & \underline{0.608} & \underline{0.697} & \underline{0.712} & \underline{24.34} & \underline{0.328} & \underline{0.718} & \underline{0.743} \\
\cellcolor{1st} \textbf{Ours}$^{\dagger}$ & \cellcolor{1st}\textbf{0.651} & \cellcolor{1st}\textbf{22.46} & \cellcolor{1st}\textbf{0.421} & \cellcolor{1st}\textbf{0.553} & \cellcolor{1st}\textbf{0.672} & \cellcolor{1st}\textbf{0.698} & \cellcolor{1st}\textbf{24.24} & \cellcolor{1st}\textbf{0.355} & \cellcolor{1st}\textbf{0.677} & \cellcolor{1st}\textbf{0.724} & \cellcolor{1st}\textbf{0.745} & \cellcolor{1st}\textbf{25.88} & \cellcolor{1st}\textbf{0.293} & \cellcolor{1st}\textbf{0.777} & \cellcolor{1st}\textbf{0.773} \\
\bottomrule
\end{tabular}
}
\vspace{-0.1cm}
\end{table*}

%% file: tbls/part_recall/item.tex
\begin{table*}[!t]
\centering
\caption{\label{tab:ade20kpart_recall} Part-mask recall of rendered SVG regions on ADE20K-Part234. We uniformly sample 200 images from the \ADE~validation set. $R@t$ is the percentage of ground-truth masks whose best matching SVG region has IoU at least $t$. $\mathrm{Avg.R}$ is the arithmetic mean of the four displayed recalls, $R@.3$, $R@.5$, $R@.7$, and $R@.8$.}
\setlength{\tabcolsep}{1.6pt}
\resizebox{0.99\textwidth}{!}{%
\begin{tabular}{ll|ccccc|ccccc|ccccc}
\toprule
\textbf{Protocol} & \textbf{Method} & \multicolumn{5}{c|}{\textbf{576 points / 64 curves}} & \multicolumn{5}{c|}{\textbf{1152 points / 128 curves}} & \multicolumn{5}{c}{\textbf{2304 points / 256 curves}} \\
& & $R@.3^{\uparrow}$ & $R@.5^{\uparrow}$ & $R@.7^{\uparrow}$ & $R@.8^{\uparrow}$ & $\mathrm{Avg.R}^{\uparrow}$ & $R@.3^{\uparrow}$ & $R@.5^{\uparrow}$ & $R@.7^{\uparrow}$ & $R@.8^{\uparrow}$ & $\mathrm{Avg.R}^{\uparrow}$ & $R@.3^{\uparrow}$ & $R@.5^{\uparrow}$ & $R@.7^{\uparrow}$ & $R@.8^{\uparrow}$ & $\mathrm{Avg.R}^{\uparrow}$ \\
\midrule
\multirow{5}{*}{Same points} & \LIVE & 42.4 & 17.2 & 5.2 & 2.7 & 16.9 & 51.3 & 20.5 & 4.7 & 1.9 & 19.6 & 56.9 & 23.3 & 6.8 & 2.8 & 22.4 \\
 & \DiffVG & 43.1 & 13.0 & 2.8 & 0.7 & 14.9 & 46.7 & 13.6 & 2.8 & 0.8 & 16.0 & 49.5 & 12.6 & 2.1 & 0.7 & 16.2 \\
 & \LIVSS & \underline{49.9} & \underline{33.4} & \underline{19.7} & \underline{12.2} & \underline{28.8} & \underline{65.4} & \underline{51.7} & \underline{30.7} & \underline{17.8} & \underline{41.4} & \underline{79.3} & \underline{61.8} & \underline{35.4} & \underline{20.1} & \underline{49.1} \\
 & SGLIVE~\cite{zhou2024segmentation} & 42.8 & 18.4 & 6.2 & 3.5 & 17.7 & 50.5 & 21.9 & 6.2 & 2.9 & 20.4 & 57.4 & 23.4 & 5.6 & 2.7 & 22.3 \\
 & \cellcolor{1st} \textbf{Ours} & \cellcolor{1st}\textbf{75.1} & \cellcolor{1st}\textbf{60.5} & \cellcolor{1st}\textbf{37.2} & \cellcolor{1st}\textbf{21.1} & \cellcolor{1st}\textbf{48.5} & \cellcolor{1st}\textbf{85.1} & \cellcolor{1st}\textbf{70.7} & \cellcolor{1st}\textbf{44.0} & \cellcolor{1st}\textbf{26.3} & \cellcolor{1st}\textbf{56.5} & \cellcolor{1st}\textbf{94.3} & \cellcolor{1st}\textbf{79.3} & \cellcolor{1st}\textbf{51.3} & \cellcolor{1st}\textbf{31.8} & \cellcolor{1st}\textbf{64.2} \\
\midrule
\multirow{2}{*}{Same paths} & \LIVSS & \underline{60.6} & \underline{45.5} & \underline{28.7} & \underline{18.2} & \underline{38.3} & \underline{74.5} & \underline{56.6} & \underline{33.5} & \underline{19.4} & \underline{46.0} & \underline{85.1} & \underline{68.0} & \underline{38.7} & \underline{22.9} & \underline{53.7} \\
& \cellcolor{1st} \textbf{Ours} & \cellcolor{1st}\textbf{77.7} & \cellcolor{1st}\textbf{64.4} & \cellcolor{1st}\textbf{40.2} & \cellcolor{1st}\textbf{23.7} & \cellcolor{1st}\textbf{51.5} & \cellcolor{1st}\textbf{85.8} & \cellcolor{1st}\textbf{72.2} & \cellcolor{1st}\textbf{47.5} & \cellcolor{1st}\textbf{28.7} & \cellcolor{1st}\textbf{58.5} & \cellcolor{1st}\textbf{93.9} & \cellcolor{1st}\textbf{79.3} & \cellcolor{1st}\textbf{54.7} & \cellcolor{1st}\textbf{34.8} & \cellcolor{1st}\textbf{65.7} \\
\bottomrule
\end{tabular}
}
\vspace{-0.1cm}
\end{table*}

%% file: tbls/coco_comparison/item.tex
\begin{table*}[!t]
\centering
\caption{\label{tab:coco_point} Quantitative evaluation on COCO-AOKVQA under the \emph{Same points} protocol, including reconstruction and image/text semantic metrics.}
\setlength{\tabcolsep}{2.5pt}
\resizebox{0.99\textwidth}{!}{%
\begin{tabular}{l|ccccc|ccccc|ccccc}
\toprule
\textbf{Method}
& \multicolumn{5}{c|}{\textbf{576 points}}
& \multicolumn{5}{c|}{\textbf{1152 points}}
& \multicolumn{5}{c}{\textbf{2304 points}} \\
& $PSNR^{\uparrow}$ & $\text{S-XLIP}^{\uparrow}$ & $CLIP^{\uparrow}$ & $\text{S-XLIP}_{t2i}^{\uparrow}$ & $CLIP_{t2i}^{\uparrow}$
& $PSNR^{\uparrow}$ & $\text{S-XLIP}^{\uparrow}$ & $CLIP^{\uparrow}$ & $\text{S-XLIP}_{t2i}^{\uparrow}$ & $CLIP_{t2i}^{\uparrow}$
& $PSNR^{\uparrow}$ & $\text{S-XLIP}^{\uparrow}$ & $CLIP^{\uparrow}$ & $\text{S-XLIP}_{t2i}^{\uparrow}$ & $CLIP_{t2i}^{\uparrow}$ \\
\midrule
\LIVE
& 19.76 & \underline{0.371} & \underline{0.607} & \underline{0.234} & \underline{0.238}
& 21.46 & \underline{0.534} & \underline{0.675} & \underline{0.291} & \underline{0.261}
& \underline{23.41} & \underline{0.692} & \underline{0.752} & \underline{0.331} & \underline{0.276} \\

\LIVSS
& 18.29 & 0.349 & 0.599 & 0.226 & 0.236
& 19.94 & 0.472 & 0.653 & 0.273 & 0.255
& 22.16 & 0.641 & 0.724 & 0.323 & 0.272 \\

\DiffVG
& \underline{20.35} & 0.271 & 0.591 & 0.188 & 0.228
& \underline{21.63} & 0.387 & 0.642 & 0.233 & 0.249
& 22.95 & 0.559 & 0.717 & 0.294 & 0.270 \\

\cellcolor{1st} \textbf{Ours}
& \cellcolor{1st}\textbf{20.53} & \cellcolor{1st}\textbf{0.460} & \cellcolor{1st}\textbf{0.644} & \cellcolor{1st}\textbf{0.270} & \cellcolor{1st}\textbf{0.254}
& \cellcolor{1st}\textbf{22.27} & \cellcolor{1st}\textbf{0.601} & \cellcolor{1st}\textbf{0.703} & \cellcolor{1st}\textbf{0.313} & \cellcolor{1st}\textbf{0.268}
& \cellcolor{1st}\textbf{23.78} & \cellcolor{1st}\textbf{0.724} & \cellcolor{1st}\textbf{0.760} & \cellcolor{1st}\textbf{0.341} & \cellcolor{1st}\textbf{0.279} \\
\bottomrule
\end{tabular}
}
\vspace{-0.3cm}
\end{table*}

%% file: figs/comparison/item.tex
\begin{figure*}
\vspace{-1em}
  \centering
  \includegraphics[width=\linewidth]{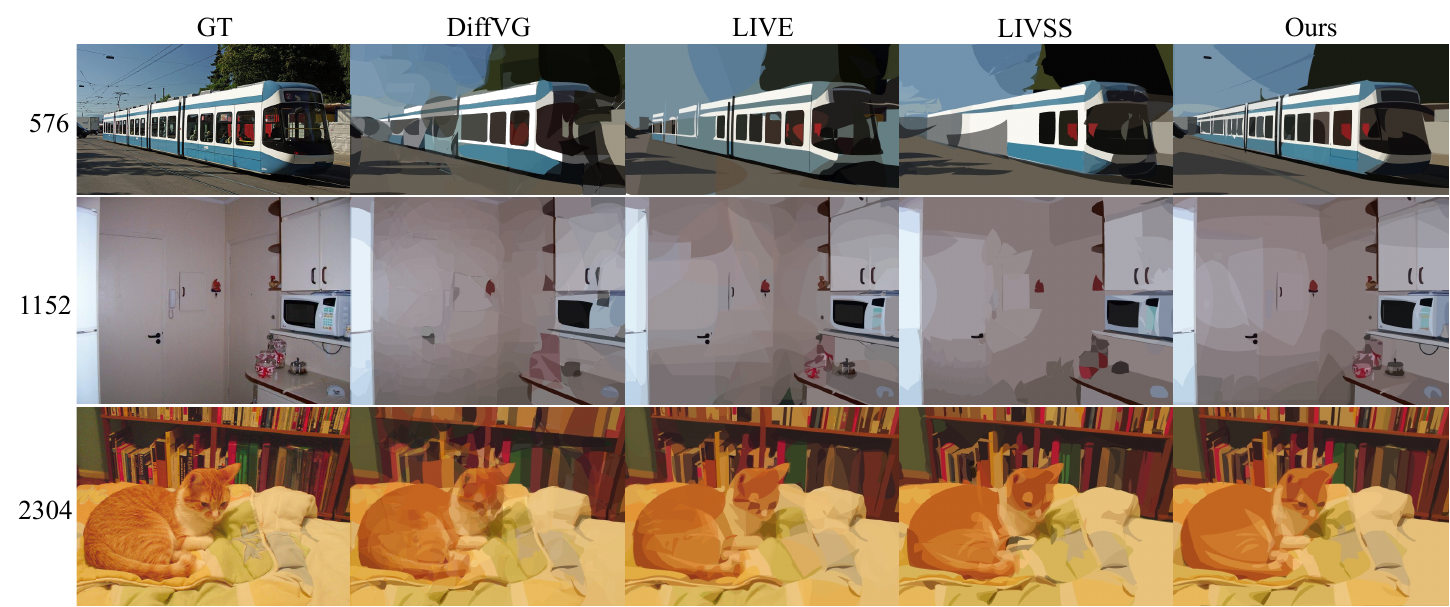}
\caption{
 \textbf{Image vectorization comparison.}
We compare our method with \DiffVG, \LIVE, and \LIVSS. Our SVGs achieve better reconstruction fidelity and semantic alignment with the ground-truth images, preserving object boundaries and fine-grained structural details.
}
\label{fig:main-comparison}
\end{figure*}

%% file: tbls/hierarchy_quality/item.tex
\begin{table}[H]
\vspace{-0.1cm}
\centering
\caption{Hierarchy and semantic quality on ADE20K. Results are percentages (higher is better).}
\label{tab:hierarchy_quality}
\small
\setlength{\tabcolsep}{5pt}
\begin{tabular}{c|ccc}
\toprule
\textbf{Threshold} & Child Containment & Tagging Reliability & Labeled Subpart Recall \\
\midrule
0.3 & 98.13 & 82.56 & 65.98 \\
0.5 & 95.55 & 68.25 & 47.06 \\
0.7 & 86.65 & 46.62 & 27.00 \\
\bottomrule
\end{tabular}
\vspace{-0.2cm}
\end{table}

%% file: tbls/tagging_ablation/item.tex
\begin{table}[!t]
\centering
\caption{
\textbf{Ablation of tree-guided semantic assignment.}
We report exact accuracy for masked-label prediction on ADE20K-Part nodes.
\emph{Full Img.} denotes the original input image; \emph{Crop Img.} denotes target-region crops; \emph{Obj./Par.} denotes object and parent semantic labels; \emph{Sib./Child} denotes sibling and child semantic labels; \emph{Geo.} denotes parent, sibling, and child geometric layout; and \emph{Color} denotes SVG region color statistics.
}
\vspace{-.5em}
\small
\resizebox{0.7\columnwidth}{!}{%
\begin{tabular}{lcccccc|c}
\toprule
ID
& Full Img.
& Crop Img.
& Obj./Par.
& Sib./Child
& Geo.
& Color
& Acc. \\
\midrule
A0 & \cmark & \cmark &        &        &        &        & 58.5 \\
A1 & \cmark & \cmark & \cmark &        &        &        & 67.7 \\
A2 & \cmark & \cmark & \cmark & \cmark &        &        & 68.9 \\
A3 & \cmark & \cmark & \cmark & \cmark & \cmark &        & 80.2 \\
\rowcolor{oursblue}
A4 & \cmark & \cmark & \cmark & \cmark & \cmark & \cmark & \textbf{81.3} \\
\midrule
\multicolumn{8}{l}{\textit{Negative controls}} \\
A5 & \cmark &        & \cmark & \cmark & \cmark & \cmark & 69.2 \\
A6 &        &        & \cmark & \cmark & \cmark & \cmark & 60.7 \\
\bottomrule
\end{tabular}
}
\label{tab:qwen-subpart-fair-ablation}
\end{table}

%% file: figs/tagging/item.tex
\begin{figure*}
\vspace{-.3em}
  \centering
    \includegraphics[width=\linewidth]{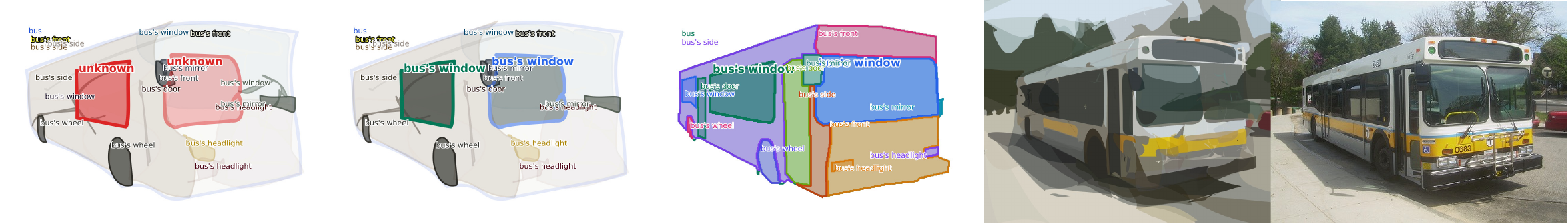}
  \includegraphics[width=\linewidth]{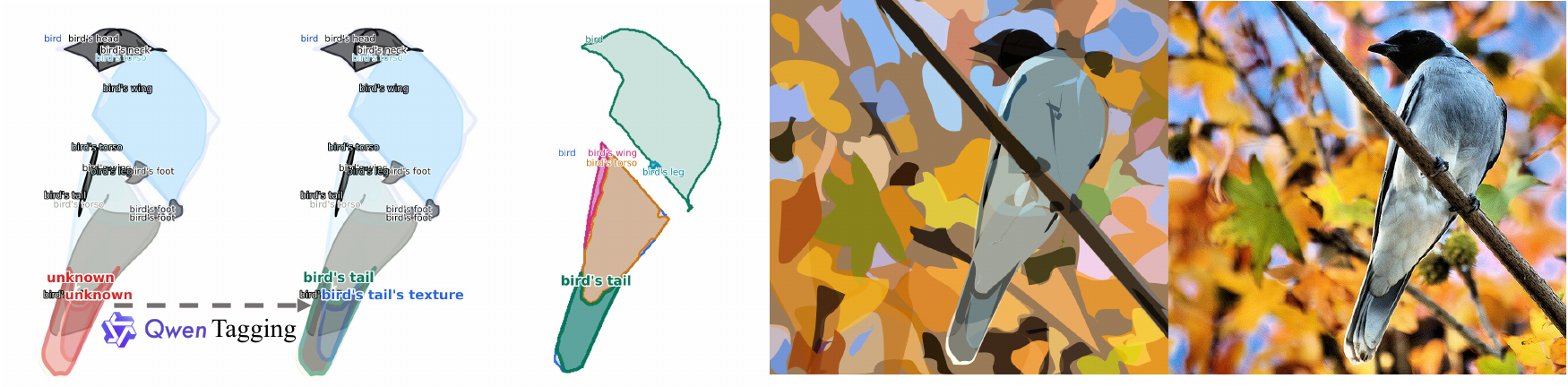}
\caption{\textbf{Semantic SVG tagging.}
We leverage the tree context and Qwen3.5-27B to predict the unknown tags of SVGs shown on the left. For clearer visualization, we only render SVG components whose semantic tags belong to the target object. The results show that \Ours~enables fine-grained semantic reasoning, where the tail region is predicted as \textit{``bird tail''}, while a smaller sub-region is further recognized as the \textit{``texture''} of the bird tail.
}
\label{fig:tagging}
\vspace{-.5em}
\end{figure*}

%% file: tbls/ablation_priors/item.tex
\begin{table}[t]
\centering
\caption{Segmentation priors and vectorization design on ADE20K with 2304 points. The first two rows compare segmentation priors within the LIVSS pipeline; the last two rows compare LIVSS and Ours using the same Semantic-SAM and Grounded-SAM priors.}
\label{tab:segmentation_prior_control}
\small
\setlength{\tabcolsep}{4pt}
\resizebox{0.95\linewidth}{!}{%
\begin{tabular}{ll|ccccc}
\toprule
\textbf{Method} & \textbf{Segmentation priors} & $SSIM^{\uparrow}$ & $PSNR^{\uparrow}$ & $LPIPS^{\downarrow}$ & $\text{S-XLIP}^{\uparrow}$ & $CLIP^{\uparrow}$ \\
\midrule
LIVSS & SAM & 0.707 & 24.12 & 0.336 & 0.695 & 0.734 \\
LIVSS & Semantic-SAM + Grounded-SAM & 0.707 & 24.23 & 0.338 & 0.708 & 0.735 \\
\rowcolor{oursblue}
\cellcolor{1st} \textbf{Ours} & Semantic-SAM + Grounded-SAM & \textbf{0.746} & \textbf{25.90} & \textbf{0.291} & \textbf{0.775} & \textbf{0.777} \\
\bottomrule
\end{tabular}%
}
\vspace{-1.0em}
\end{table}

%% file: tbls/ablation_tree/item.tex
\begin{table}[t]
\centering
\caption{Tree-related ablations on ADE20K with 2304 points using Semantic-SAM and Grounded-SAM priors. Depth denotes depth-adaptive simplification; Tree Pruning denotes tree-guided pruning and merging. Checkmarks indicate enabled components. Tree pruning requires the tree structure, so removing the tree also disables tree pruning. Ours is the same final model as in Table~\ref{tab:ablation_v4_2304}.}
\label{tab:tree_isolation}
\small
\setlength{\tabcolsep}{4pt}
\resizebox{0.95\linewidth}{!}{%
\begin{tabular}{lccc|ccccc}
\toprule
\textbf{Method} & \textbf{Depth} & \textbf{Tree} & \shortstack{\textbf{Tree}\\\textbf{Pruning}} & $SSIM^{\uparrow}$ & $PSNR^{\uparrow}$ & $LPIPS^{\downarrow}$ & $\text{S-XLIP}^{\uparrow}$ & $CLIP^{\uparrow}$ \\
\midrule
w/o depth simplification & $\times$ & \cmark & \cmark & 0.717 & 24.84 & 0.332 & 0.714 & 0.744 \\
w/o tree and tree pruning & \cmark & $\times$ & $\times$ & 0.708 & 24.29 & 0.337 & 0.705 & 0.737 \\
w/o Tree Prune & \cmark & \cmark & $\times$ & 0.743 & 25.71 & 0.296 & 0.769 & 0.770 \\
\rowcolor{oursblue}
Ours & \cmark & \cmark & \cmark & \textbf{0.746} & \textbf{25.90} & \textbf{0.291} & \textbf{0.775} & \textbf{0.777} \\
\bottomrule
\end{tabular}%
}
\end{table}

%% file: tbls/ablation_modules/item.tex
\begin{table}[!t]
\centering
\caption{\label{tab:ablation_v4_2304} Ablation of L1-SAM-Prompting and segmentation backbone on ADE20K with 2304 points.}
\vspace{-.5em}
\setlength{\tabcolsep}{3pt}
\resizebox{0.75\columnwidth}{!}{%
\begin{tabular}{l|ccccc}
\toprule
\textbf{Method} & $SSIM^{\uparrow}$ & $PSNR^{\uparrow}$ & $LPIPS^{\downarrow}$ & $\text{S-XLIP}^{\uparrow}$ & $CLIP^{\uparrow}$ \\
\midrule

w/o L1-SAM-Prompting & 0.740~{\scriptsize(-0.006)} & 25.41~{\scriptsize(-0.49)} & 0.298~{\scriptsize(+0.007)} & 0.762~{\scriptsize(-0.013)} & 0.770~{\scriptsize(-0.007)} \\

w/ UnSAMv2 & 0.741~{\scriptsize(-0.005)} & 25.73~{\scriptsize(-0.17)} & 0.297~{\scriptsize(+0.006)} & 0.763~{\scriptsize(-0.012)} & 0.769~{\scriptsize(-0.008)} \\

\rowcolor{oursblue}
Ours & \textbf{0.746} & \textbf{25.90} & \textbf{0.291} & \textbf{0.775} & \textbf{0.777} \\
\bottomrule
\end{tabular}
}
\end{table}

%% file: sec/5_conclusion.tex
\section{Conclusion}
We present Lang-SVG, a tree-structured SVG representation for hierarchical image vectorization and semantic labeling. Lang-SVG organizes SVG primitives into semantic object-part hierarchies and incorporates depth-adaptive simplification, Tree-Guided Pruning and Merging, and L1-SAM-Prompting to reduce redundancy, recover missing regions, and improve rendering fidelity. We further introduce a tree-guided semantic tagging method that leverages hierarchical context to guide MLLMs in labeling SVG nodes. To evaluate these representations beyond reconstruction fidelity, we propose a new evaluation protocol that measures structural quality and semantic alignment. Experiments on ADE20K demonstrate that Lang-SVG achieves state-of-the-art performance in both rendering fidelity and structural quality, providing structured and semantically labeled SVG representations for editing and downstream tasks.

\paragraph{Limitations and Future Work.}
Our method relies on open-vocabulary segmentation models to provide initial labels and language labels, such that the parsing quality is still partially bounded by the accuracy and granularity of these upstream models \cite{li2026seeing}. Missing parts, ambiguous masks, or incomplete labels may be inherited by the SVG tree, especially for small subparts and texture-like regions. Although our tree-guided MLLM assignment can recover many unknown tags using structural context, it is not yet specifically trained for SVG semantic understanding.
A promising future direction is to fine-tune MLLMs for SVG-aware labeling. By training on SVG trees, rendered vector regions, parent-child relations, and part-level annotations, such models could assign more accurate semantic tags to vector components, reduce dependence on open-vocabulary segmentation, and improve fine-grained parsing of complex natural images.

%% file: sec/6_acknowledgments.tex
\section*{Acknowledgments}
This research was supported in part by the eBay eRUPT program and the SC NASA EPSCoR Research Grant Program (No.~21-117-RGP-CU-011). This research used in part resources on the Palmetto Cluster at Clemson University under NSF awards MRI 1228312, II NEW 1405767, MRI 1725573, and MRI 2018069. JL is supported by DNRF grant P1. The views expressed in this article do not necessarily represent the views of any funders.

%% file: sec/7_appendix.tex
\newpage
\section{Appendix}

This appendix provides an additional evaluation on SVG-style images, the hierarchical SVG construction algorithms, the ADE20K tagging system prompt, and further qualitative vectorization and semantic-tagging results.

\subsection{Generalization to SVG-Style Images}
\label{sec:appendix_iconfont}
We evaluate 40 complex SVG-style images selected from Iconfont to complement the natural-image benchmarks. All methods use the same budget of 576 control points. Table~\ref{tab:iconfont} shows that \Ours~achieves the highest PSNR, SSIM, and CLIP similarity and the lowest LPIPS. Compared with the strongest PSNR baseline, SGLIVE, PSNR increases from 28.62 to 29.28. LIVE achieves the highest StructXLIP score, while our method scores 0.9549. These results demonstrate that the pipeline also applies to SVG-style content. This experiment measures reconstruction and perceptual quality; hierarchy and semantic-label quality are evaluated separately using ADE20K object--part annotations in Sec.~\ref{sec:hierarchy_quality}.
\input{tbls/iconfont/item}

\input{figs/svg_style/item}

\subsection{Algorithms for Hierarchical SVG Construction}
\label{sec:appendix_algorithms}
Algorithms~\ref{alg:svg_pipeline} and~\ref{alg:update_tree} detail the overall pipeline and the tree update procedure. Open-vocabulary masks initialize labeled semantic branches, while multi-granularity masks and reconstruction-driven proposals add finer nodes. For each new mask, candidate parents are selected by spatial containment; the deepest candidate is preferred, with IoU used to break ties. The full-image root provides a fallback. This geometric assignment rule supports masks at different granularities, although imperfect segmentation and partially overlapping regions can still produce ambiguous relations. The hierarchy evaluation in Sec.~\ref{sec:hierarchy_quality} measures the resulting parent alignment and label consistency.

\begin{algorithm}[H]
\caption{Overall pipeline for hierarchical SVG construction}
\label{alg:svg_pipeline}
\begin{algorithmic}[1]
\Require Image $I$; open-vocabulary masks $\mathcal O$ with labels and hierarchy $\mathcal H_{\mathcal O}$; multi-granularity masks $\mathcal M$; point budget $B$; thresholds $\tau_{\mathrm{parent}},\tau_{\mathrm{prune}}$
\Ensure Semantic SVG tree $\mathcal T$ and flattened SVG $S$
\State Create root $r$ with the full-image background mask
\State Initialize $\mathcal T$ from $\mathcal O$, preserving labels and $\mathcal H_{\mathcal O}$
\State Simplify and vectorize initial masks according to their tree depths
\State $\mathcal T\gets\Call{UpdateTree}{\mathcal T,\mathcal M,I,B,\tau_{\mathrm{parent}},\tau_{\mathrm{prune}}}$
\While{the point budget has not been reached}
    \State Compute the reconstruction residual between $I$ and $\operatorname{Render}(\mathcal T)$
    \State $\mathcal M_{\mathrm{new}}\gets$ generate masks using L1-SAM-Prompting
    \If{$\mathcal M_{\mathrm{new}}=\emptyset$}
        \State $\mathcal M_{\mathrm{new}}\gets$ generate masks from high-residual regions
    \EndIf
    \If{$\mathcal M_{\mathrm{new}}=\emptyset$}
        \State \textbf{break}
    \EndIf
    \State $\mathcal T\gets\Call{UpdateTree}{\mathcal T,\mathcal M_{\mathrm{new}},I,B,\tau_{\mathrm{parent}},\tau_{\mathrm{prune}}}$
    \If{no further update can be admitted under the point budget}
        \State \textbf{break}
    \EndIf
\EndWhile
\State Label unknown nodes using the visual input, tree context, geometry, and color
\State Flatten $\mathcal T$ into painter's rendering order to obtain $S$
\State Store semantic labels and tree relations as SVG metadata
\State \Return $\mathcal T,S$
\end{algorithmic}
\end{algorithm}

\begin{algorithm}[H]
\caption{\textsc{UpdateTree}: mask assignment, optimization, and pruning}
\label{alg:update_tree}
\begin{algorithmic}[1]
\Require Current tree $\mathcal T$; candidate masks $\mathcal Q$; image $I$; point budget $B$; thresholds $\tau_{\mathrm{parent}},\tau_{\mathrm{prune}}$
\Ensure Updated tree $\mathcal T$
\Function{UpdateTree}{$\mathcal T,\mathcal Q,I,B,\tau_{\mathrm{parent}},\tau_{\mathrm{prune}}$}
    \For{each mask $q\in\mathcal Q$, from coarse to fine}
        \State $\mathcal P\gets\{v\in\mathcal T:\,|q\cap m_v|/|q|\geq\tau_{\mathrm{parent}}\}$
        \If{$\mathcal P=\emptyset$}
            \State $p\gets r$
        \Else
            \State $d^*\gets\max_{v\in\mathcal P}\operatorname{depth}(v)$
            \State $\mathcal P^*\gets\{v\in\mathcal P:\operatorname{depth}(v)=d^*\}$
            \State $p\gets\arg\max_{v\in\mathcal P^*}\operatorname{IoU}(q,m_v)$
        \EndIf
        \State Attach $q$ as a child of $p$, using \emph{unknown} for any missing tag
        \State Simplify and vectorize $q$ according to its tree depth
    \EndFor
    \State Optimize SVG paths and colors against $I$
    \State Prune or merge redundant parent--child and sibling nodes within the same semantic branch using color, overlap, and $\tau_{\mathrm{prune}}$
    \State Preserve semantically labeled nodes and enforce the point budget $B$
    \State \Return $\mathcal T$
\EndFunction
\end{algorithmic}
\end{algorithm}
\clearpage
\subsection{ADE20K Tagging System Prompt}
\label{sec:appendix_tagging_prompts}
\label{sec:tagging_ade_prompt}
We provide the system prompt used for closed-set ADE20K semantic tag recovery below. The complete set of 234 candidate labels is supplied as \texttt{candidate\_labels} in the user message. Only the final JSON output schema is omitted from this display.

\begingroup
\lstset{
  basicstyle=\ttfamily\fontsize{7.5}{9}\selectfont,
  columns=fullflexible, keepspaces=true,
  breaklines=true, breakatwhitespace=false,
  showstringspaces=false, showspaces=false, showtabs=false,
  frame=single, framerule=0.4pt, framesep=6pt,
  rulecolor=\color{PaperRule}, backgroundcolor=\color{PaperCanvas},
  numbers=none, tabsize=2,
  xleftmargin=7pt, xrightmargin=7pt,
  aboveskip=0.7em, belowskip=0.7em
}
\begin{lstlisting}
You are an ADE20KPart234 scene-graph part classification evaluator.

You will receive:
- the original full image
- a crop around the containing object, when available
- a crop around the target bbox, when available
- metadata copied from a layered scene-graph JSON
- known object/part context and sibling layout from that JSON
- bbox layout metadata for the target, parent, siblings, and parent children
- a closed candidate list of valid part/subpart labels

The ground-truth target node tag has been hidden. Classify the semantic part or subpart
inside the target bbox using the image and the unmasked scene-graph context.

Rules:
- Choose exactly one label from candidate_labels.
- Use the target bbox metadata to locate the masked part/subpart inside the full image and object crop.
- Use the object crop for object-level context, not as the target region itself.
- Use the target bbox crop for local visual details.
- Use ancestor_tags, parent_tag, sibling_tags, child_tags, parent_layout,
  sibling_bboxes, and parent_child_layout from the JSON as context.
- candidate_labels are the only valid outputs. They are full hierarchical labels;
  reason about their leaf part names, but output the exact candidate label string.
- predicted_label must exactly match one candidate label string.
- Do not invent a new label outside the candidate list.
- Return exactly one JSON object.
- Do not output prose, markdown, or reasoning text.

Before choosing the label, silently run this parent-to-subpart checklist:
1. Identify the parent object/part from ancestor_tags, parent_tag, and parent_layout.
2. Generate plausible child parts for that parent, then eliminate candidates whose
   position, size, color/material, or sibling layout is inconsistent.
3. Compare the target bbox with every sibling bbox under the same parent. Existing
   known siblings are strong constraints; masked siblings may still provide layout
   evidence through their bbox and color but not their hidden tag.
4. Use bbox_average_rgb and visual color/material continuity as supporting evidence.
   Similar color/texture can indicate another segment of the same object; very
   different color may indicate clothing, shoe, material detail, or texture.
5. Avoid impossible duplicate anatomy or object parts unless the image and layout
   clearly support duplication. Choose "unknown" only if no candidate label is
   spatially and visually plausible, but prefer a valid candidate when evidence is
   sufficient.

Examples of the intended reasoning:
- If parent is person and known siblings already include left/right arm, a third
  arm candidate at the lower body is usually wrong; a lower bbox near feet is more
  likely leg, foot, shoe, or clothing depending on shape and color.
- If parent is person's leg and the target bbox is at the bottom end with a shoe-like
  color/shape, prefer the candidate whose leaf is foot or shoe over hand or arm.
- If parent is person's face/head, small upper symmetric boxes are likely eyes; a
  central box below eyes and above mouth is likely nose; a lower face box is more
  likely mouth/chin than eye.
- If parent is car/vehicle, lower left/right boxes are likely wheel/tire; boxes in
  the upper body are more likely window, door, mirror, or light.
- If one known sibling is a wheel on the lower left and the target is a similar
  colored/size lower-right box, the target is likely another wheel candidate.
\end{lstlisting}
\endgroup
\clearpage

\subsection{Additional Qualitative Results}
We compare the SVG outputs generated by \DiffVG, \LIVE, \LIVSS, and our method under the same point-budget setting. These examples show that \Ours~better preserves the visual structure of natural images while maintaining high rendering fidelity. We also include more semantic SVG tagging examples, where our tree-guided semantic assignment predicts meaningful object, part, and subpart labels for SVG components.

\input{figs/more_comparison/item}

\input{figs/more_comparison_low_budget/item}

\input{figs/more_tagging/item}

%% file: tbls/iconfont/item.tex
\begin{table}[htbp]
\centering
\caption{Vectorization of 40 complex SVG-style images selected from Iconfont ~\cite{iconfont} under the same 576-point budget. Best results are bold.}
\label{tab:iconfont}
\small
\setlength{\tabcolsep}{5pt}
\begin{tabular}{l|ccccc}
\toprule
\textbf{Method} & $SSIM^{\uparrow}$ & $PSNR^{\uparrow}$ & $LPIPS^{\downarrow}$ & $\text{S-XLIP}^{\uparrow}$ & $CLIP^{\uparrow}$ \\
\midrule
DiffVG & 0.9125 & 26.46 & 0.1274 & 0.8899 & 0.9397 \\
LIVE & 0.9349 & 28.19 & 0.0882 & \textbf{0.9616} & 0.9704 \\
SGLIVE & 0.9308 & 28.62 & 0.0815 & 0.9546 & 0.9728 \\
LIVSS & 0.9319 & 28.24 & 0.0945 & 0.9491 & 0.9696 \\
\rowcolor{oursblue}
Ours & \textbf{0.9384} & \textbf{29.28} & \textbf{0.0782} & 0.9549 & \textbf{0.9766} \\
\bottomrule
\end{tabular}
\end{table}

%% file: figs/svg_style/item.tex
\begin{figure*}[h]
\vspace{-0.5em}
  \centering
  \includegraphics[width=\linewidth]{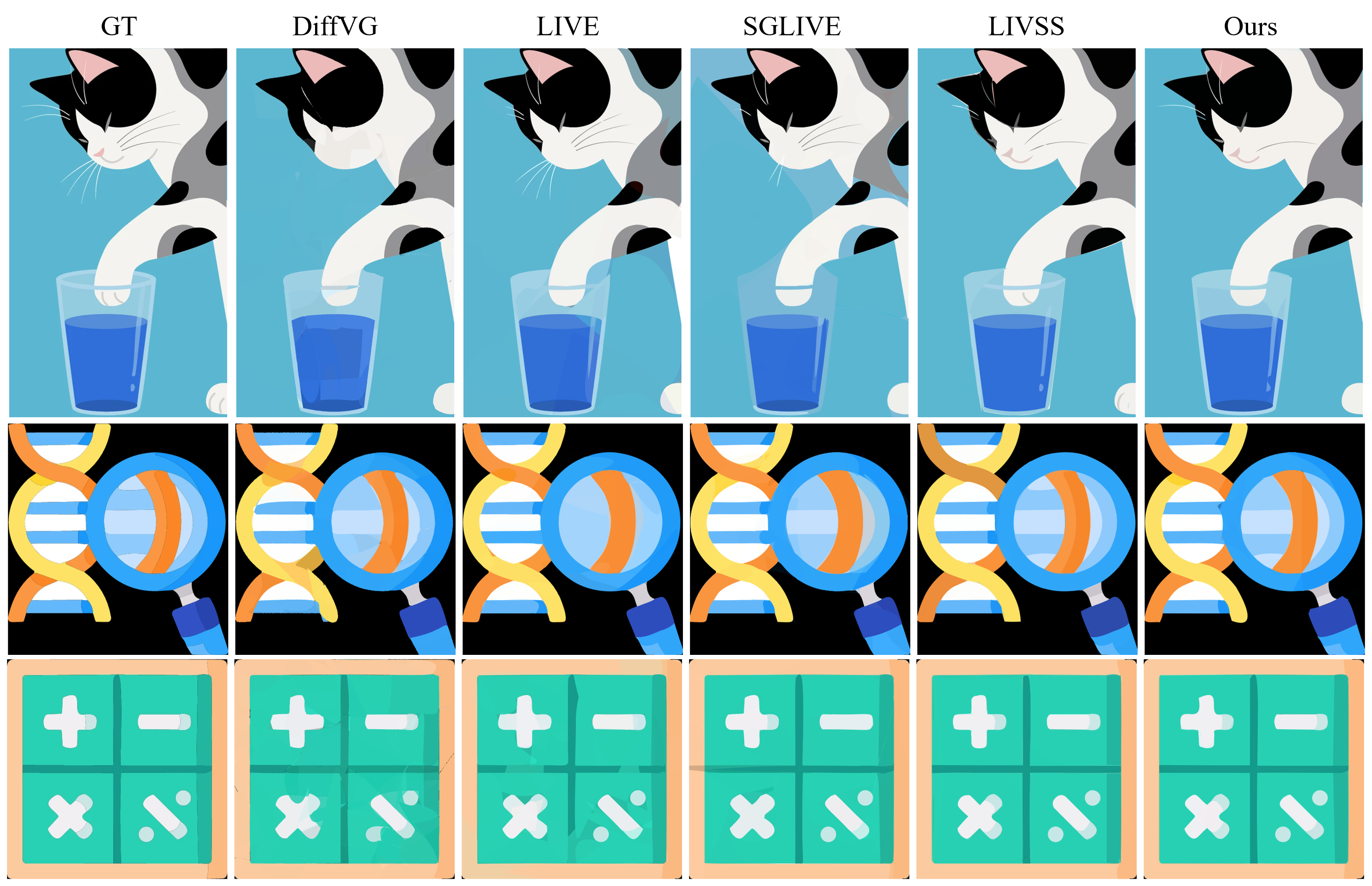}
\caption{
\textbf{SVG-style qualitative comparison.}
We compare SVGs produced by \DiffVG, \LIVE, SGLIVE~\cite{zhou2024segmentation}, \LIVSS, and our method.
}
\label{fig:svg_style_comparison}
\vspace{-0.8em}
\end{figure*}

%% file: figs/more_comparison/item.tex
\begin{figure*}[h]
\vspace{-0.5em}
  \centering
  \includegraphics[width=\linewidth]{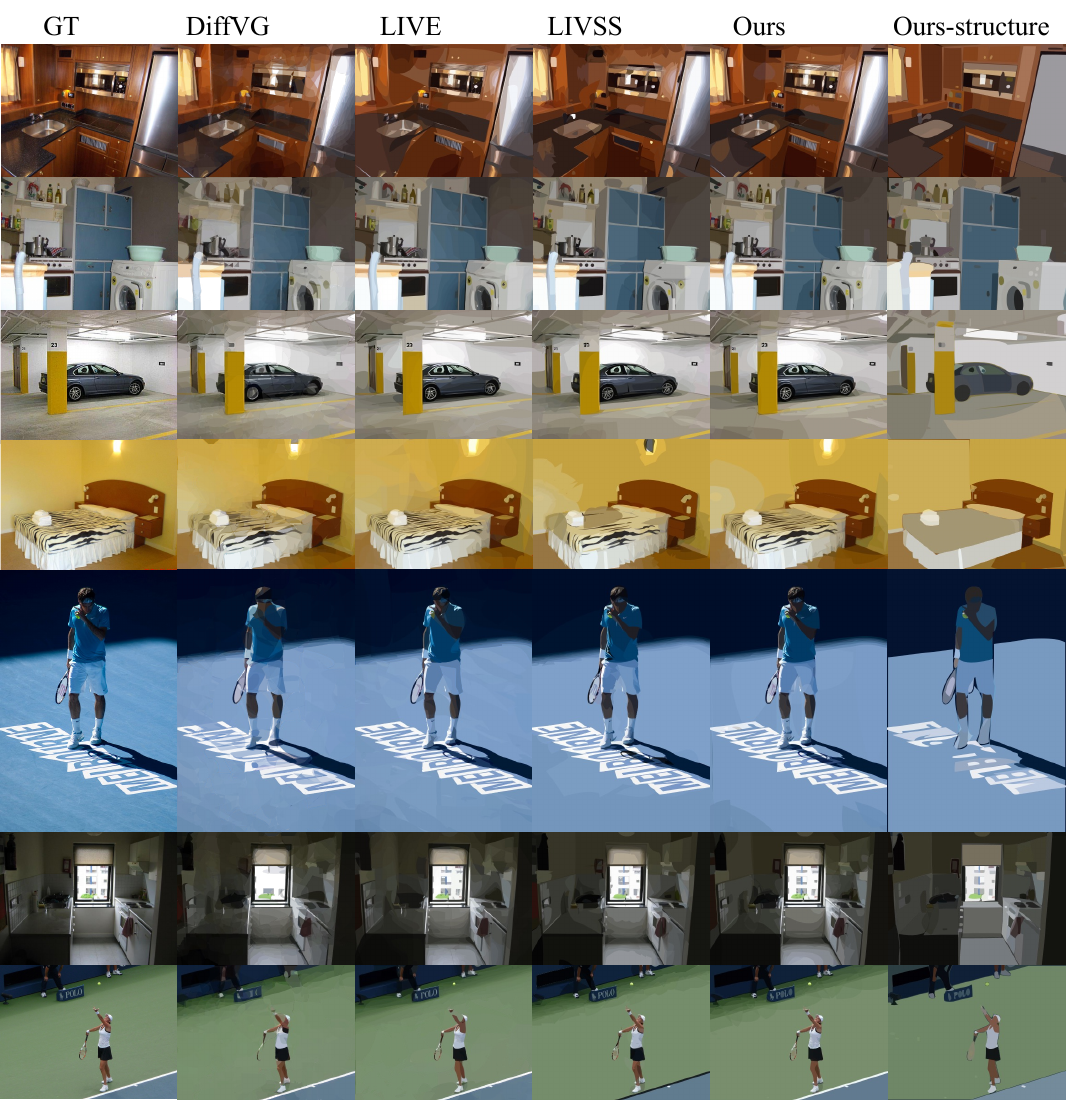}
\caption{
\textbf{More comparison results.}
We compare SVGs produced by \DiffVG, \LIVE, \LIVSS, and our method. Our SVGs better align with the target images and preserve more coherent visual structures.
}
\label{fig:more_comparison}
\vspace{-0.8em}
\end{figure*}

%% file: figs/more_comparison_low_budget/item.tex
\begin{figure*}[t]
\vspace{-0.5em}
  \centering
  \includegraphics[width=\linewidth]{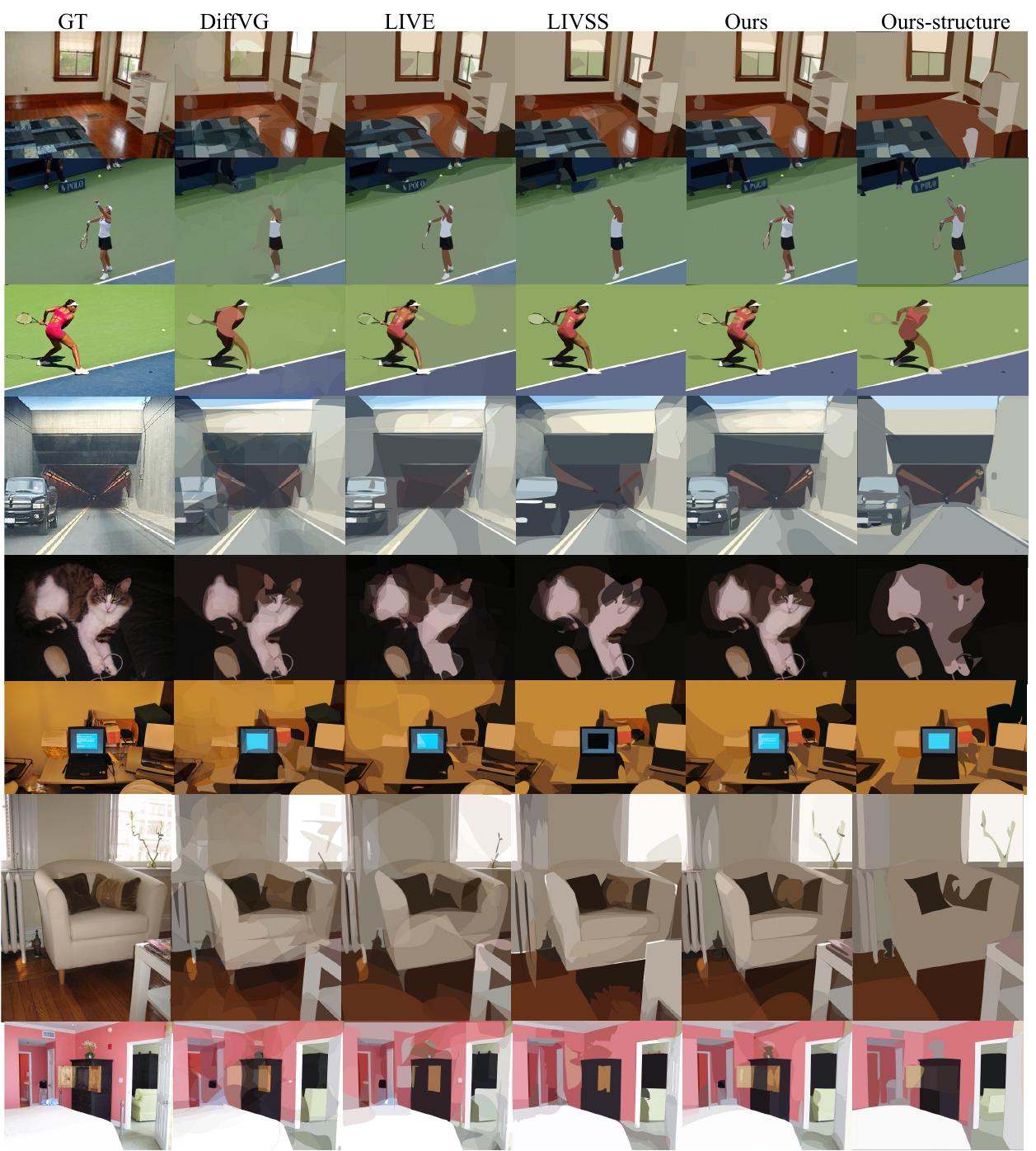}
\caption{
\textbf{More comparison results under a lower point budget.}
We compare SVGs produced by \DiffVG, \LIVE, \LIVSS, and our method. Our method produces SVGs with better visual fidelity and more coherent structure under the same point budget.
}
\label{fig:more_comparison_2}
\vspace{-0.8em}
\end{figure*}

%% file: figs/more_tagging/item.tex
\begin{figure*}[t]
\vspace{-0.5em}
  \centering
  \includegraphics[width=\linewidth]{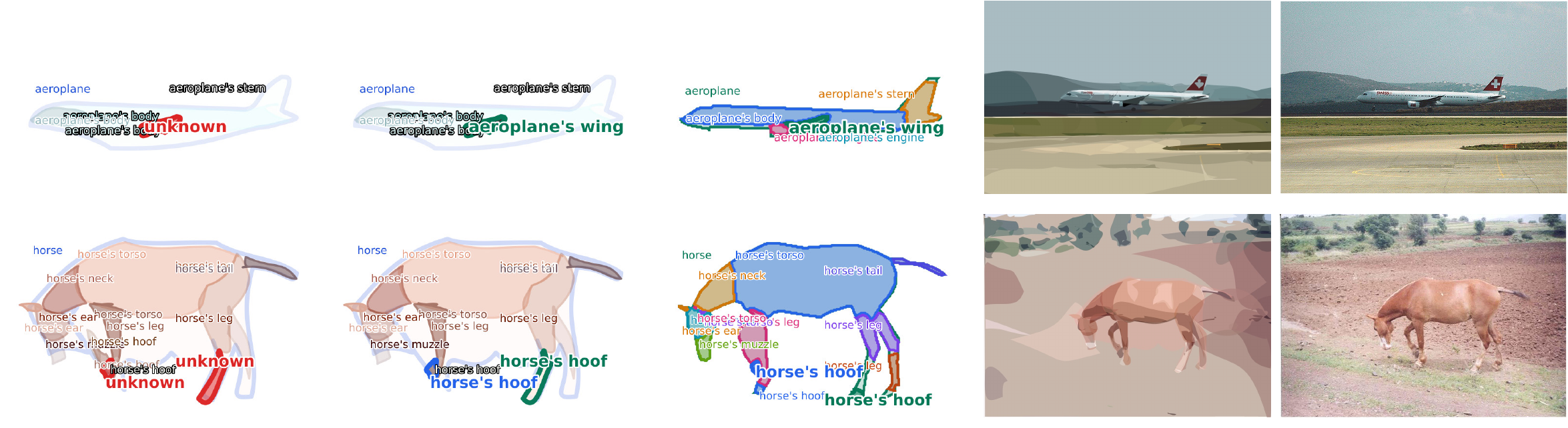}
\caption{
\textbf{More semantic SVG tagging results.}
We show additional examples of semantic tags predicted for SVG components using our tree-guided semantic assignment.
}
\label{fig:more_tagging}
\vspace{-0.8em}
\end{figure*}